\documentclass[final]{cvpr}

\usepackage{times}
\usepackage{epsfig}
\usepackage{graphicx}
\usepackage{amsmath}
\usepackage{amssymb}

\usepackage{bm,color,layout}

\newcommand{\Tref}[1]{Table~\ref{#1}}
\newcommand{\eref}[1]{Eq.~(\ref{#1})}

\newcommand{\fref}[1]{Fig.~\ref{#1}}
\newcommand{\Fref}[1]{Figure.~\ref{#1}}
\newcommand{\sref}[1]{Sec.~\ref{#1}}

\def\ie{\emph{i.e}\onedot} 

\def\etal{\emph{et al}\onedot}

\newcommand{\threesixty}{360$^{\circ}$\xspace}
\newcommand{\envlight}{near-field environment light\xspace}

\newcommand{\ENVLIGHT}{Near-field Environment Light\xspace}

\usepackage[pagebackref=true,breaklinks=true,colorlinks,bookmarks=false]{hyperref}

\def\cvprPaperID{2744} 
\def\confYear{CVPR 2021}

\begin{document}

\title{Lighting, Reflectance and Geometry Estimation from \threesixty Panoramic Stereo}




\author{\hspace{10mm}  Junxuan Li$^{1,2}$\hspace{27mm} 
Hongdong Li$^1$\hspace{23mm} 
Yasuyuki Matsushita$^3$\\
$^{1}$Australian National University\hspace{7mm} 
$^{2}$Data61-CSIRO,  Australia\hspace{3mm} 
$^{3}$Osaka University,  Japan \\
{\tt\small \{junxuan.li; hongdong.li\}@anu.edu.au}  \ \ \ \ \ \  {\tt\small yasumat@ist.osaka-u.ac.jp}
}

\maketitle

\begin{abstract}
We propose a method for estimating high-definition spatially-varying lighting, reflectance, and geometry of a scene from \threesixty stereo images. Our model takes advantage of the \threesixty input to observe the entire scene with geometric detail, then jointly estimates the scene's properties with physical constraints. We first reconstruct a \envlight for predicting the lighting at any 3D location within the scene. Then we present a deep learning model that leverages the stereo information to infer the reflectance and surface normal. Lastly, we incorporate the physical constraints between lighting and geometry to refine the reflectance of the scene. Both quantitative and qualitative experiments show that our method, benefiting from the \threesixty observation of the scene, outperforms prior state-of-the-art methods and enables more augmented reality applications such as mirror-objects insertion. 
\end{abstract}


\section{Introduction}

Intrinsic decomposition of scene properties is a long-standing and essential task in computer vision.  It includes the estimation of lighting, geometry, and reflectance of an arbitrary scene. Inferring the above properties of a scene enables us to develop various novel applications, especially in augmented reality, such as object insertion and scene modification. It is a challenging and extremely under-constrained problem because of the complexity of light transportation on complicated geometry and various material reflectances in real-world. 
The majority of previous methods used perspective cameras for solving this problem. 
However, the limited field of view of a perspective camera results in the lack of observation of the entire scene, making this inverse problem even more intractable. 

\begin{figure}
    \centering
    \includegraphics[width=0.48\textwidth]{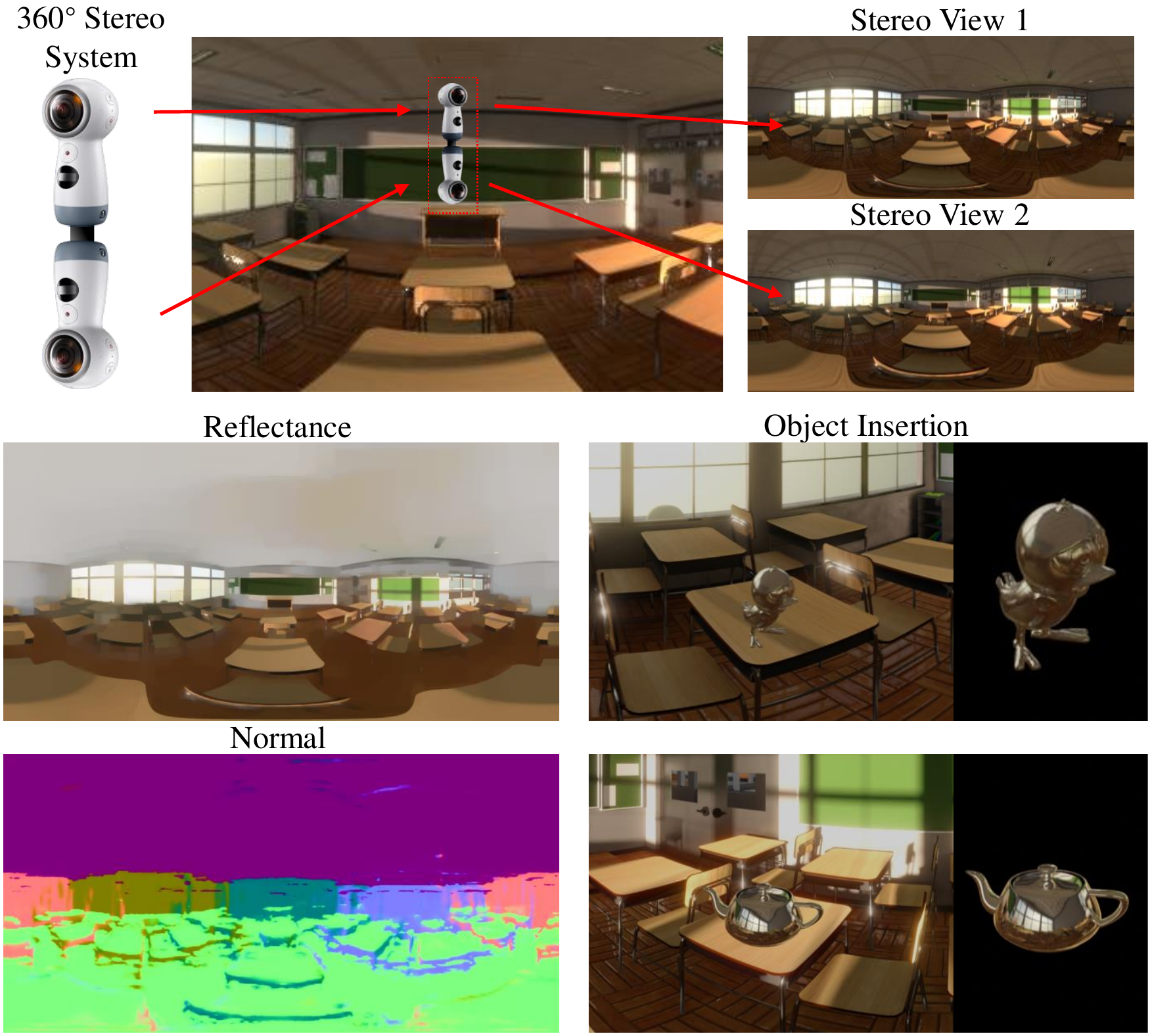}
    \caption{Our system consists of two \threesixty cameras in a top-bottom setting. We present the predicted reflectance and surface normal of the scene at the bottom-left of this figure. The bottom-right are virtual mirror-objects relighted by our illumination map at two different locations. 
    Our method can recognize the geometric difference among 3D locations and preserve high-frequency information of the lighting, enabling us to insert mirror-objects around the entire scene with appealing reflection effects. 
    }
    \label{fig:teaser}
\end{figure}

To overcome the problem, we propose a method that uses a pair of \threesixty images under equirectangular projection as input. 
Our method utilizes this input to bring up many advantages that the perspective approach does not. Firstly, the \threesixty image captures the entire scene at once, offering us an adequate observation for lighting estimation. Secondly, the stereo input naturally encodes the depth information, making the geometry estimation possible. Furthermore, by jointly leveraging the physical constraints between lighting and geometry, the reflectance can be revealed. 
\Fref{fig:teaser} illustrates the camera setting for capturing \threesixty stereo input and the estimated results of our method.

Leveraging \threesixty stereo input, we achieve two strengths in lighting estimation: \textit{(i)}
our lighting is spatially-varying and 3D coherent, which means the lighting will be different and changing smoothly for different 3D locations condition on the scene geometry;
\textit{(ii)} our lighting is in high-definition, which means it is generated in high-resolution and contains  high-frequency details of the scene to enable mirror-like objects insertion.
The lighting estimated by perspective methods rarely has these properties.
They either estimate the lighting globally~\cite{lombardi2015reflectance,legendre2019deeplight} or per-pixel-individually~\cite{song2019neural,garon2019fast,li2020inverse}, having no consistency between different locations. 
In addition, because of the limited field of view in perspective images, prior works have difficulties in `inferring' the unseen regions of the scene in high-definition. But,  our method naturally avoids this problem. 

The reflectance estimation is also an ill-posed problem under the perspective cameras~\cite{bell2014intrinsic}. 
However, with the \threesixty stereo images, the input contains more information about the scene's lighting and geometry, giving us substantial leads and more constraints to infer both reflectance and normal.


In this paper, we present a method that utilizes the strengths of the \threesixty input to jointly estimate the high-definition and spatially-varying lighting, reflectance, and geometry of the entire scene. Our contributions are:
\begin{enumerate}
    \item A \envlight that can generate spatially-varying and 3D coherent high-definition illumination maps when given any 3D location within the scene. 
    \item A deep learning model that can estimate the reflectance and surface normal of the entire scene.
    \item A rendering and refinement model that leverages the physical constraints between lighting and geometry to jointly estimate a finer reflectance.
\end{enumerate}

\section{Related Work}

\paragraph{Lighting Estimation}
Inferring the lighting of the environment enables us to render a synthetic object into real-world. 
Debevec~\cite{debevec2008rendering} used a light probe to measure an HDR illumination map.
Though a  mirror-ball is accurate in lighting estimation, it is not suitable for estimating lighting at different locations. 
Many recent studies prefer spatially-varying lighting for indoor scenarios, which predicts different lighting given different locations within the scene. \cite{gardner2017learning,gardner2019deep} use deep learning models to explicitly predict the location and intensity of the primary light sources; 
\cite{garon2019fast,zhao2020pointar} adopt the spherical harmonics lighting model for fast estimation;
\cite{xing2018automatic} assumes indoor objects located in a six-faces-box; 
\cite{srinivasan2020lighthouse} uses a deep model to represent the scene in lighting volume; 
\cite{song2019neural} warps the seen scene points into the target illumination map based on geometry estimation, then per-pixel-independently completes the unseen region by a neural network. 

Due to the limited field of view of perspective images, all the previous methods either use simplified lighting models; or simplify environment models; or hallucinate the unobserved scene's geometry and appearance, leading to lost or inconsistent lighting. It is also why previous works are unsuitable for inserting mirror-objects, which requires high-definition illumination maps and detailed lighting from the scene. 


\paragraph{Intrinsic Image Decomposition}
The studies in the intrinsic image decomposition can be divided into two folds by its input types: the object-scale and the scene-scale. An object-scale intrinsic decomposition method usually assumes global illumination. The problem can be solved using carefully designed handcrafted priors~\cite{barron2014shape}, or recently developed deep learning models~\cite{shi2017learning}.

For scene-scale decomposition problem, Barron~\etal~\cite{barron2013intrinsic} takes RGB-D as input to estimate the reflectance, lighting and normal by applying handcrafted priors. 
\cite{bi20151} proposes an $\ell_1$ norm for constraint the reflectance to be piecewise flattening. 
The prevailing deep learning methods also show its effectiveness in this task:
\cite{narihira2015direct} proposes a CNN trained by a synthetic dataset. 
The subsequent studies~\cite{fan2018revisiting,li2018cgintrinsics,zhou2019glosh} enlarge the training datasets and enrich the designs of network architectures and loss functions.
Li~\etal~\cite{li2020inverse} proposes a framework that jointly reasons shape, lighting and SVBRDF from a single perspective image. However, they simplify the lighting model and fail to consider geometry constraint between different locations within the same scene. 
Our method, taking the \threesixty input to fully observe the lighting and geometry of the entire scene, estimates the scene-level reflectance, normal, and lighting with physical constraints.  


\paragraph{\threesixty Panoramic Imaging}
Many studies focus on geometry estimation by \threesixty panoramic images. Li~\etal~\cite{li2004stereo} captures the \threesixty stereo by fixing and rotating two concentric cameras for depth estimation.
Kim and Hilton~\cite{kim20133d} proposes a 3D mesh modeling method using multiple pairs of spherical images captured by a line scan camera at different locations.
Recently, consumer-level \threesixty cameras are used for depth estimation from the video clip~\cite{im2016all} and a stereo-pair~\cite{wang2020360sd}. Other than depth estimation, Banterle~\etal~\cite{banterle2013envydepth} takes an annotated high dynamic range (HDR) panoramic environment map for local illumination recovery. Our work fills a literature gap by estimating the lighting, geometry, and reflectance from the \threesixty stereo input.


\section{Method Overview}

\begin{figure*}
	\centering
	\includegraphics[width=0.95\textwidth]{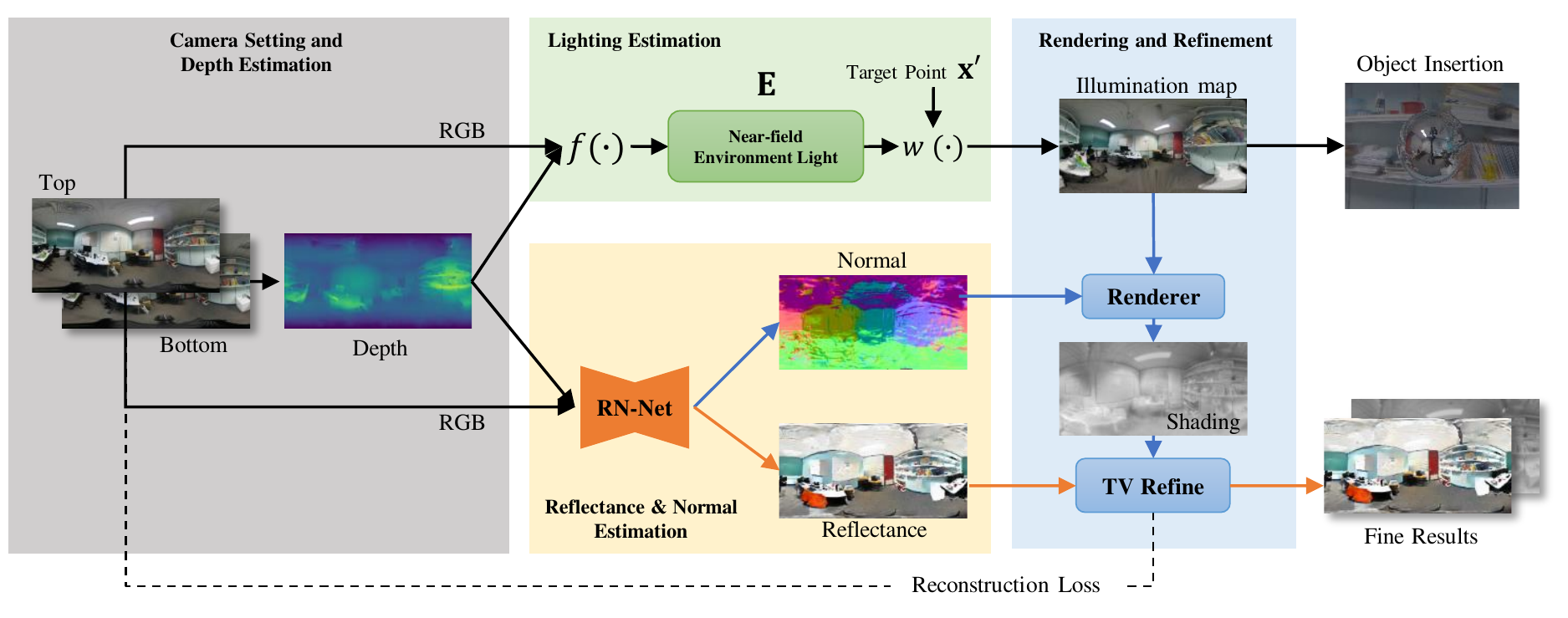}
	\caption{\textbf{System overview}. 
	We first estimate the depth of a stereo \threesixty input. Then a \envlight is reconstructed from the input images and estimated depth for estimating illumination map later. We parallelly apply an RN-Net on inputs and estimated depth for reflectance and normal estimation. Finally, we use the illumination map and normal map to render the shading, which then jointly refines the reflectances.
	}
	\label{fig:overview}
\end{figure*}

Our method uses two \threesixty images taken from a \threesixty stereo camera to estimate the target scene's lighting, geometry, and reflectance. It consists of four modules, as illustrated in \fref{fig:overview}.
The first module shows our \threesixty camera setting and depth estimation from the stereo input (see \fref{fig:overview} gray part, \sref{sec:camera}).
The second module is the lighting estimation that computes a \envlight from the input images and estimated depth. We treat each scene point as a light source in the 3D space and build a \envlight. With this \envlight, given any 3D location within the scene, our method can reconstruct the corresponding illumination map in high-definition for object insertion and relighting (see \fref{fig:overview} green part, \sref{sec:light}).      
The third module is reflectance and normal estimation. It is a deep learning model, named RN-Net, for estimating the reflectance map and surface normal of the entire scene. To preserve the high-frequency information from the input, we take the input with the resolution to be $512\times 1024$. To tackle the large input size, we proposed a pyramid structure for RN-Net to estimate the reflectance and surface normal from small to large (see \fref{fig:overview} orange part, \sref{sec:reflectance}). 

With these three modules, our method can obtain the scene's lighting, reflectance, and geometry at a certain granularity. To obtain more refined estimates of shadings and reflectances, we use the fourth module that performs physically-based rendering and refinement that aims at minimizing the reconstruction loss with the input (see \sref{sec:rendering} blue part).

\section{Proposed Method}
\subsection{Camera Setting and Depth Estimate}\label{sec:camera}
\begin{figure}
	\centering
	\includegraphics[width=0.45\textwidth]{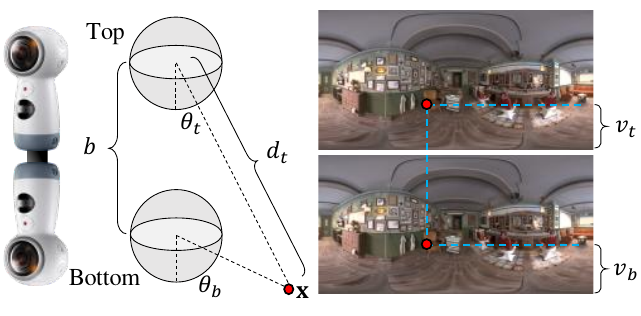}
	\caption{The system comprises two Samsung Gear \threesixty cameras. 
	Once calibrated, a point $\mathbf{x}$ in the scene will be aligned in the \threesixty images only with a vertical angular disparity $\Delta\theta$.}
	\label{fig:two_cameras}
\end{figure}

Our imaging setup consists of two \threesixty cameras in a top-bottom setting as shown in \fref{fig:two_cameras}. A similar setup has also been used by \cite{wang2020360sd}.
This top-bottom arrangement ensures only the vertical disparities between two \threesixty images. 
The captured \threesixty image is in equirectangular projection, allowing us to capture the entire scene at once, while the stereo images enable us to estimate the geometry at a low cost.

As illustrated in \fref{fig:two_cameras}, for a point $\mathbf{x} = [x, y, z]^\top \in \mathbb{R}^3$ in the 3D space, let its  projection on the top and bottom images be $\mathbf{u}_t = [u_t,v_t]^\top \in \mathbb{R}^2$
and $\mathbf{u}_b = [u_b, v_b]^\top \in \mathbb{R}^2$,
respectively. When the two cameras are aligned vertically with the same $v$-axis, the displacement $\Delta\theta \in \mathbb{R}$ of $\mathbf{x}$ on two images can be given by
\begin{align}
    \Delta\theta = \theta_b - \theta_t  = \frac{\pi}{h} \left(v_b-v_t\right).
    \label{eq:angular_displacement}
\end{align}
$h$ is the height of image, $v_b$ and $v_t$ are the $v$-coordinates of the projected image points $\mathbf{u}_t$ and $\mathbf{u}_b$ respectively. 
The distance between $\mathbf{x}$ and the top camera, $d_t$ is given by triangulation as:
\begin{align}
    d_t = b  \left( \frac{\sin{\theta_t}}{\tan{\Delta\theta}} +\cos{\theta_t} \right) ,
    \label{eq:d_t}
\end{align}
where $b \in \mathbb{R}_+$ is the baseline between the two cameras. 

Therefore, with a stereo matching method to find matchings along the vertical direction, we can obtain the given point's angular disparity and depth by the above Eqs.~(\ref{eq:angular_displacement}) and (\ref{eq:d_t}). In this paper, we use a recent CNN-based stereo method, 360SD-Net~\cite{wang2020360sd} for depth estimation.

\subsection{Lighting Estimation}\label{sec:light}
To preserve the high-frequency lighting and geometry information of the \threesixty environment, we propose the second module to reconstruct a \envlight for estimating the high-definition spatially-varying illumination map given an arbitrary 3D location in the scene. 

\paragraph{\ENVLIGHT}
We assume that all the observed scene materials are diffuse when estimating the lighting.
It is also a convention, and a good approximation, to treat scene material as diffuse when doing lighting estimation~\cite{xing2018automatic,zhou2019glosh}. 

For simplicity, we use the top camera as the reference camera to omit the subscript of the notations.
Following the notation above, $(\mathbf{c}, \mathbf{u}, d)$ represents a pixel in the \threesixty image, where $\mathbf{c} \in \mathbb{R}^3$ is the RGB observation of a point in the camera, \ie, the pixel value with three channels;
and $\mathbf{u}$ is its position on the reference camera with coordinates $[u,v]^T$; 
$d \in \mathbb{R}_+$ is the estimated depth from the previous step.

We define  $f(\cdot)$ as the projection function between a pixel in \threesixty image under the equirectangular projection to the world coordinates:
$ (\mathbf{c}, \mathbf{x}) = f(\mathbf{c}, \mathbf{u}, d). $
Applying the projection $f(\cdot)$ to all pixels on the \threesixty image will give us the representation of the \envlight $\mathbf{E} \in \mathbb{R}^{n\times 6}$, where $n$ is the number of points, as:
\begin{align}
    \mathbf{E} = \left\{(\mathbf{c}_i, \mathbf{x}_i) \; | \; i \in \text{pixels}\right\}.
\end{align}
Each point in the scene is treated as a light source with intensity $\mathbf{c}$ and its position $\mathbf{x}$.

\paragraph{Illumination Map}

Given an arbitrary 3D point $\mathbf{x}'$ in the scene, we re-project the \envlight $\mathbf{E}$ to the new point to generate an illumination map for $\mathbf{x}'$ by: 
\begin{align}
    \left\{  (\mathbf{c}'_i, \mathbf{u}'_i, d'_i) \; | \; i \in \text{pixels}   \right\}  = g (\mathbf{E},  \mathbf{x}'),
\end{align}
where $g(\cdot)$ projects the coordinates from 3D to 2D illumination map. 
However, due to the sparsity of the \envlight, the reconstructed illumination map contains pixels without any projection of the lights, while some pixels may have many lights fall into. 
Hence, we need to refine the illumination map to sort out empty and overlapped pixels. 
The refinement function $r(\cdot)$ is defined as that for the position $\mathbf{u}'_{i}$ with many projected lights, only select the light with the minimum depth value $d'$. 
This manner simulates the occlusion effect in the real world, where one may obstruct another light. 
We use the nearest interpolation method to extrapolate those empty pixels.
In summary, our reconstructed illumination map is given by applying functions $g(\cdot)$ and $r(\cdot)$ in sequence. To simplify, we use function $w$ to denote a composition of functions $r\circ g$.
The reconstructed illumination map $\mathbf{M}' \in \mathbb{R}^{n\times 6}$ at  position $\mathbf{x}'$ is:
\begin{align}
    \mathbf{M}' = \left\{  (\mathbf{c}'_i, \mathbf{u}'_i, d'_i) \; | \; i \in \text{pixels}   \right\} 
    & = w (\mathbf{E},  \mathbf{x}').
\end{align}

\begin{figure}
	\centering
	\includegraphics[width=0.52\textwidth]{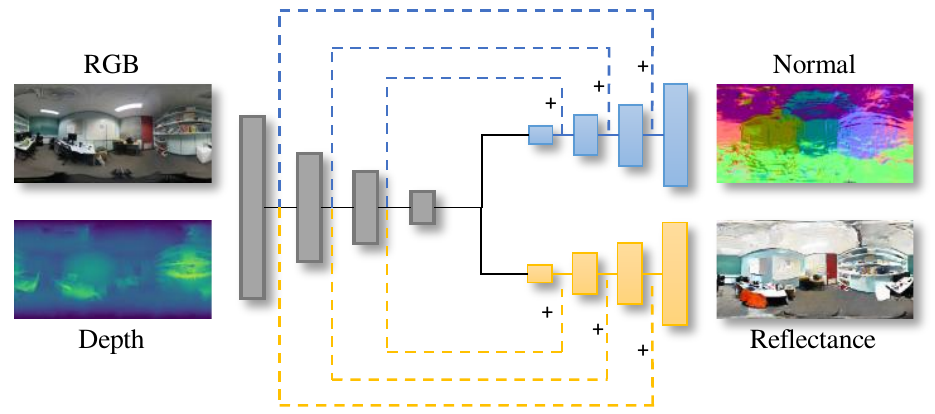}
	\caption{The architecture of RN-Net. It follows a U-Net structure. Please see supplementary material for more details.}
	\label{fig:RN_net}
\end{figure}

\subsection{Reflectance and Normal Estimation}\label{sec:reflectance}
The third module is a convolutional neural network, named RN-Net, for predicting the reflectance and surface normal of the entire scene at a large resolution. 

As shown in \fref{fig:overview}, it takes the reference \threesixty image and estimated depth  as the input to infer the reflectance and normal. \Fref{fig:RN_net} shows detailed network architecture. 
It processes the input with four encoder-blocks. 
Each block consists of two convolutional layers and a short skip connection between input and output and will down-sample the feature size into half, similar to ResNet~\cite{he2016deep}. 
Then, for normal and reflectance estimation, we apply another four decoder-blocks for each task. The decoder is similar to the encoder but will up-sample the feature size twice larger at the output. 
Besides the short skip connection within each block, we also add a long skip connection between each layer, as shown in \fref{fig:RN_net}.

To tackle the large input size of the \threesixty images, we apply a pyramid structure to the RN-Net. 
The input is first scaled to a different size, then feed into different RN-Net for training.  In the end,  we up-sample all the results to the original resolution and add them together to get the reflectance and normal estimation. In this paper, our network takes RGB image with estimated depth as the input $ \mathbf{I}_1 \in \mathbb{R}^{512\times 1024 \times 4}$ and scales it to four times smaller to be $\mathbf{I}_{\frac{1}{4}} \in \mathbb{R}^{128\times 256 \times 4}$. The overall structure can be illustrated as:
\begin{align}
\left\{
\begin{array}{lcl}
    \mathbf{R}_1, \mathbf{N}_1 &=& \Phi_1(\mathbf{I}_1), \\
    \mathbf{R}_{\frac{1}{4}}, \mathbf{N}_{\frac{1}{4}} &=& \Phi_{\frac{1}{4}} (\mathbf{I}_{\frac{1}{4}}), \\
    \mathbf{R}, \mathbf{N} &=& \text{up}(\mathbf{R}_{\frac{1}{4}})+\mathbf{R}_1, \; \text{up}(\mathbf{N}_{\frac{1}{4}})+\mathbf{N}_1, 
\end{array}
\right.
\end{align}
where $\mathbf{R}_1$, $\mathbf{N}_1$, $\mathbf{R}_{\frac{1}{4}}$, and $\mathbf{N}_{\frac{1}{4}}$ are the reflectance map and normal map at the original size and $\frac{1}{4}$ size respectively; $\Phi_1(\cdot) and \Phi_{\frac{1}{4}}(\cdot)$ are the RN-Nets in different scales; 
$\text{up} (\cdot)$ operator represents the bilinear up-sampling;  $\mathbf{R}, \mathbf{N}$ are the estimated results of this pyramid structure.

We use a scale-invariant loss for training reflectance:
\begin{align}
    \mathcal{L}_{\mathbf{R}} = \left\|s \mathbf{R} - \mathbf{R^*} \right\|_2^2 + \left\|  s \nabla  \mathbf{R}  - \nabla \mathbf{R^*} \right\|_1,
\end{align}
where $\mathbf{R^*}$ is the ground truth of the reflectance map; $s$ is a scale factor computed by applying least square regression between $\mathbf{R}$ and $\mathbf{R^*}$; $\nabla  \mathbf{R}$ is the gradient of $\mathbf{R}$. 
For the training of surface normal, we define the loss as:
\begin{align}
    \mathcal{L}_{\mathbf{N}} = - \mathbf{N}^T \mathbf{N^*} + \left\| \nabla  \mathbf{N} - \nabla  \mathbf{N^*} \right\|_1,  
\end{align}
where the $\mathbf{N^*}$ is the ground truth of surface normal. Here, the first term of $\mathcal{L}_{\mathbf{N}}$ is the cosine loss between normal, while the second is gradient loss. Both of the reflectance and normal loss adopt the gradient loss, which makes results piecewise smooth, as shown in many previous works~\cite{li2018cgintrinsics,fan2018revisiting,zhou2019glosh}. 
The total training loss is $ \mathcal{L} = \mathcal{L}_{\mathbf{R}} + \mathcal{L}_{\mathbf{N}} $.

\begin{figure*}
	\centering
	\includegraphics[width=0.91\textwidth]{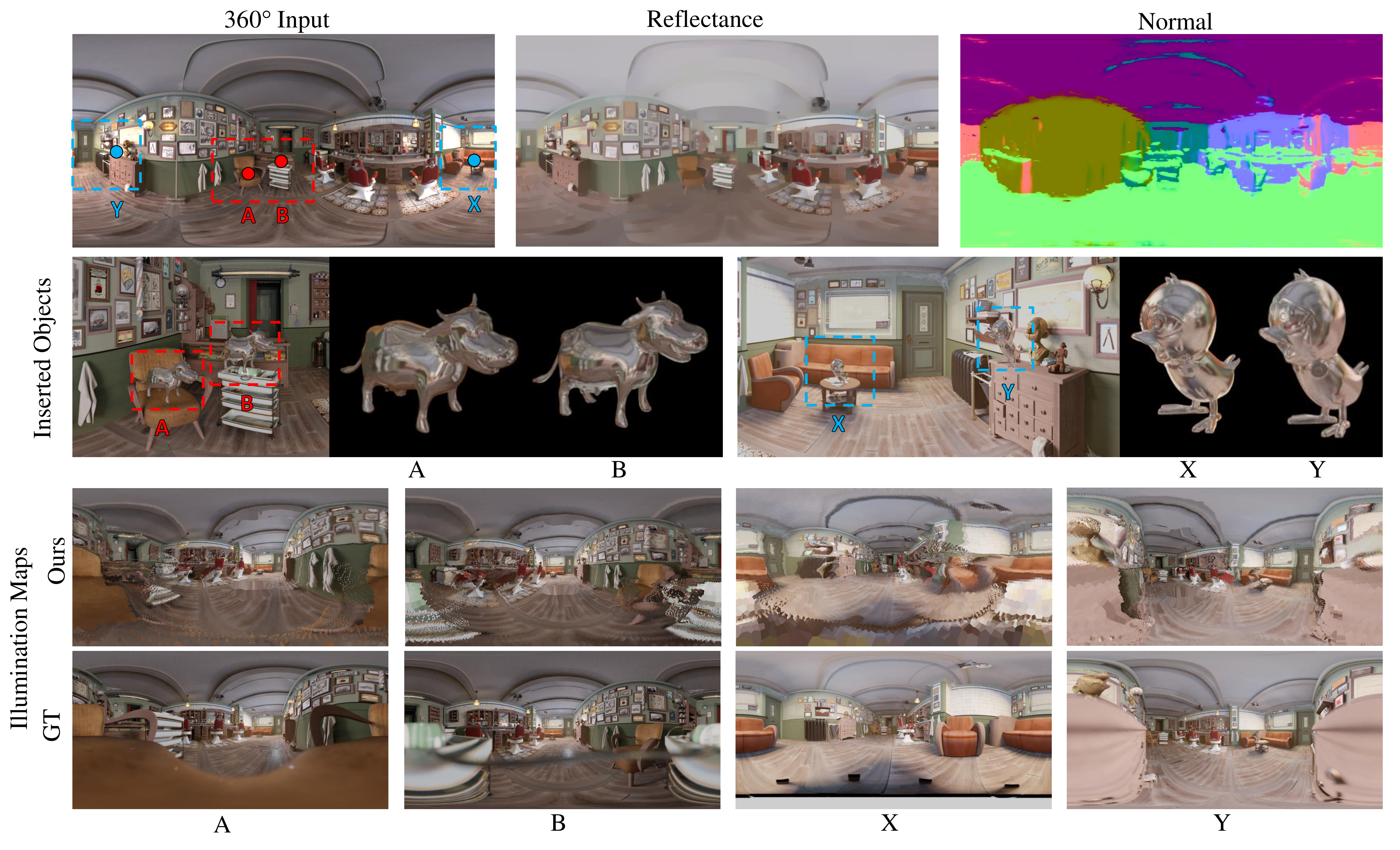}
	\vspace{-3mm}
	\caption{The \threesixty input and results from our method on the synthetic scene `barbershop.' We use our estimated illumination maps to virtually relight mirror-objects at four different locations within the scene. The last two rows present the corresponding estimated illumination maps and ground truth at each location. Please notice how the inserted `cow' at location `A' and `B' consistently reflect the surroundings, demonstrating the spatially-coherency of our lighting. The changes between the highlighted areas of the objects inserted at `X' and `Y' also indicate the variance of our lighting at different locations. By comparing with the ground truth, our method can reconstruct the illumination map with high-frequency details and accurately reflect the geometry of the scene. 
	}
	\label{fig:barbershop_results}
\end{figure*}

\subsection{Rendering and Refinement}\label{sec:rendering}
The \threesixty images observe the entire scene at once to provide lighting and geometry of the environment as constraints for solving this ill-posed reflectance estimation problem.
Our last module utilizes the estimated illumination maps and surface normal from previous steps, incorporating physical insights, to render and refine the shading and reflectance map. 

\paragraph{Shading Rendering}
For each pixel with index $i$ and its image location $\mathbf{u}_i$ in the \threesixty image $\mathbf{I}$, we first compute the corresponding 3D location $\mathbf{x}_i$ and estimate an illumination map centered at the 3D location by $\mathbf{M}_i = w(\mathbf{E}, \mathbf{x}_i)$. As we assume all the scene material to be diffuse, the $i$-th pixel's shading value $\mathbf{S}_i \in \mathbb{R}^3$  can be computed by the integration of the illumination map and dot product between light direction and surface normal.   
Hence, the $i$-th pixel's shading value is given by:
\begin{align}\label{eg:render}
    \mathbf{S}_i = \sum_{j \in \mathbf{M}_i} \mathbf{c}_j  \max \left( {\mathbf{l}_{j}}^T  \mathbf{N}_i, 0 \right), 
\end{align}
where $\mathbf{c}_{j}$ and $\mathbf{l}_{j} \in \mathbb{R}^3$  are the light intensity and light direction of light at $j$-th pixel of illumination map $\mathbf{M}_i$, respectively; and $\mathbf{N}_i \in \mathbb{R}^3$ is the estimated surface normal of the $i$-th pixel.

To render the whole shading map $\mathbf{S}\in\mathbb{R}^{512\times 1024 \times 3}$, we iterate every pixel $i$ in the input image, reconstruct the corresponding illumination map $\mathbf{M}_i$ and compute every shading value by \eref{eg:render}.

\paragraph{Refinement Using TV Regularization}
The input RGB image $\mathbf{I}$ can now be reconstructed by taking the product between shading map $\mathbf{S}$ and reflectance map $\mathbf{R}$. However, errors and noise will inevitably occur at every step of the process. Here, we refine the results from previous steps by optimizing a target energy function:
\begin{align}
    \mathcal{L}_{TV} = || \mathbf{I} - s \mathbf{R}\odot \mathbf{S} ||_{2}^2 + \lambda_1 ||\nabla \mathbf{R} ||_1  + \lambda_2 ||\nabla \mathbf{S} ||_{2}^2 ,
\end{align}
where $\odot$ denotes the Hadamard product; $s$ is a scale factor computed by applying least square regression between $\mathbf{I}$ and $\mathbf{R}\odot \mathbf{S}$. Our energy loss takes a similar form to total variation regularization~\cite{rudin1992nonlinear} and  serves in a similar task, \ie, to minimize noise and reject outliers. 

The first term can be interpreted as the reconstruction loss between the input and our reconstructed input image. The second term, applying $\ell_1$ loss on the gradient of reflectance map, aims to constrain reflectance map being piecewise smooth. While the third term, $\ell_2$ loss on the gradient of shading, aims to suppress the abrupt changes in shading effects. In this paper, we set the hyper parameters as $\lambda_1=0.1, \lambda_2=10$. 

We use the reflectance map $\mathbf{R}$ and shading map $\mathbf{S}$ from the previous step as the initialization. Then update the two maps based on the gradient descend algorithm.
\begin{align}
    \mathbf{S}' = \mathbf{S} - \gamma \dfrac{\partial \mathcal{L}_{TV} }{ \partial \mathbf{S}}, \quad
    \mathbf{R}' = \mathbf{R} - \gamma \dfrac{\partial \mathcal{L}_{TV} }{ \partial \mathbf{R}}.
\end{align}

\section{Implementation Details and Dataset}
\subsection{Training Details}
\paragraph{Training Data for RN-Net}
We use a public dataset Structured3D~\cite{zheng2019structured3d} as our training data for RN-Net. The Structured3D is a synthetic indoor dataset with  rich layouts and interior designs. It has $21835$ images of different rooms with $512\times 1024$ resolution in equirectangular projection. 
It also provides us with ground truth surface normal and reflectance map for supervised learning. The only inconsistency between this dataset and our requirements is that it does not provide stereo inputs. To overcome this, when training RN-Net,  we take the ground truth depth map and added with smoothed Gaussian noise as input to simulate the errors in depth estimation. 

\paragraph{Training of RN-Net}
Our model is trained from scratch with Adam~\cite{kingma2014adam} optimizer. We first train the small scale of RN-Net $\Phi_{\frac{1}{4}}$ with batch size to be $32$ and the learning rate to be $10^{-3}$. After $100$ epochs, we combine both RN-Nets to further train it with the large scale images. Batch size is $12$ for another $50$ epochs in the large size. The learning rate starts from $10^{-3}$, then being half every $10$ epochs. 
The RN-Net is implemented by PyTorch; trained on a single GPU NVIDIA GTX 1080Ti for around $20$ hours.

\paragraph{Total Variation Refinement} 
For optimization on $\mathcal{L}_{TV}$, we adopt Adam~\cite{kingma2014adam} optimizer with fixed learning rate $10^{-4}$. In addition, to avoid overfitting, we apply a $\ell_2$ loss between the initialization and optimized results, acting as the weight-decay.
The model is implemented in PyTorch and converges after $1000$ iterations in 2 minutes.

\subsection{Testing Data} \label{sec:data_eval}
We prepare two datasets for testing. A synthetic dataset rendered by Blender~\cite{blender}, and a real dataset captured by our \threesixty cameras.

\begin{figure*}
    \centering
    \hspace{-2mm} Li~\etal~\cite{li2020inverse} \hspace{21mm} Lighthouse~\cite{srinivasan2020lighthouse} \hspace{25mm} Ours \hspace{28mm} Ground Truth
	\includegraphics[width=0.98\textwidth]{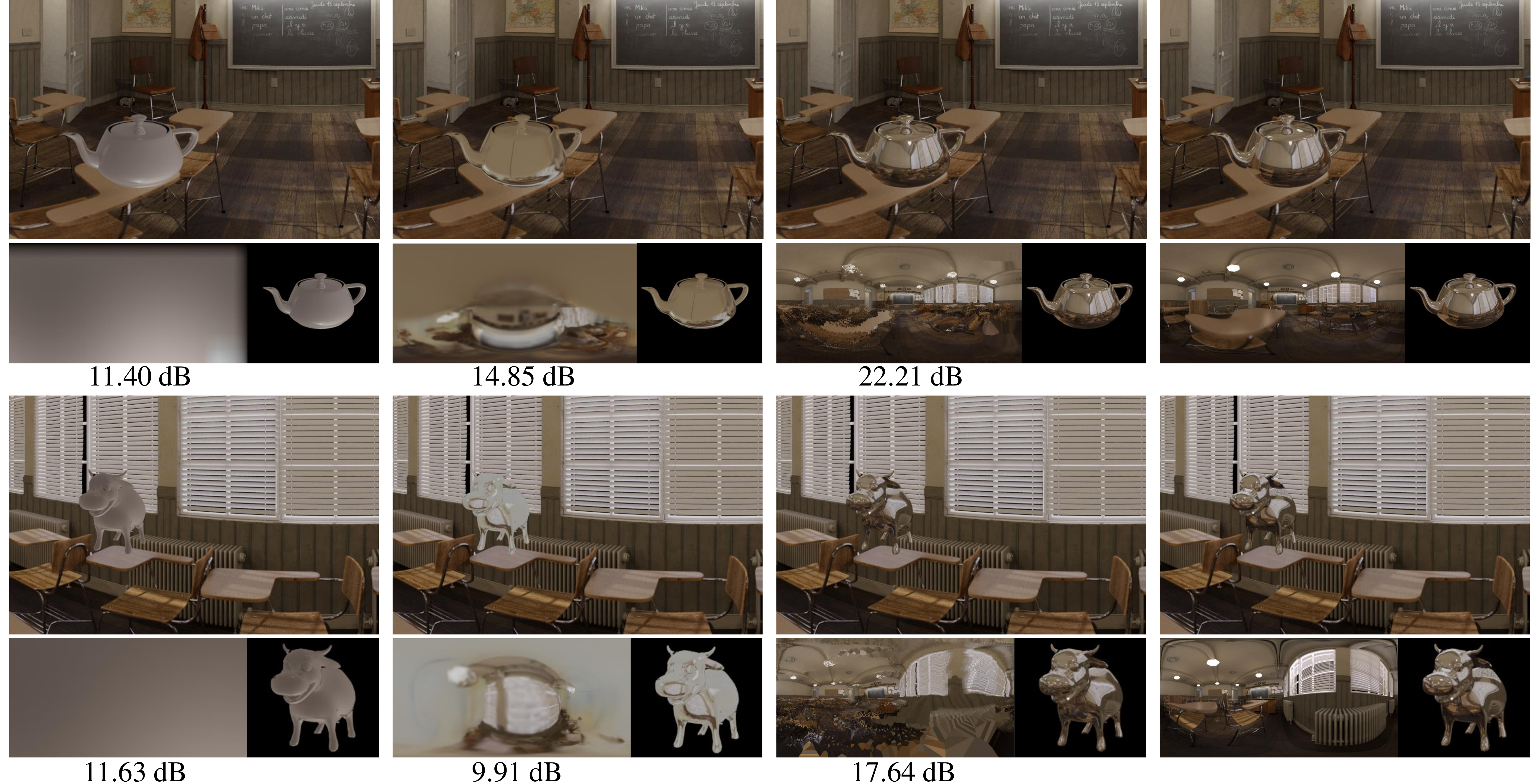}
	\caption{\textbf{Comparison to previous works} on inserted mirror-objects and estimated illumination maps. 
    The value of peak signal-to-noise ratio (higher is better) between each estimated illumination map and ground truth is shown at the bottom of each image. Our method outperforms all the previous works by providing illumination maps with rich details close to the ground truth. 
	}
	\label{fig:classroom_lighting}
\end{figure*}

\textbf{Synthetic data.}
We create a synthetic dataset by Blender to simulating the complex light transportation of an indoor scene. Each scene consists of two vertically-aligned cameras to capture the whole environment in an equirectangular projection. The renderer provides us with the ground truth of reflectance, normal, and illumination map for quantitative evaluation. We showcase some results from our synthetic scenes: `school' in \fref{fig:teaser}; `barbershop' in \fref{fig:barbershop_results}; `classroom' in \fref{fig:classroom_lighting}; and `bedroom' in \fref{fig:albedo_normal}.

\textbf{Real data.}
As shown in \fref{fig:two_cameras}, we connect and fix two Samsung Gear \threesixty cameras in a top-bottom manner. The two cameras are calibrated 
to vertically align two images and measure the baseline. We present the captured data `office' in \fref{fig:office_real}.
Following the convention on many \threesixty datasets~\cite{zheng2019structured3d,wang2020360sd}, both the real and synthetic images are captured with the input resolution to be $512 \times 1024$.

\textbf{Pre-process for Comparison to Previous Works.}
We choose two recently published works for our lighting estimation comparison~\cite{li2020inverse,srinivasan2020lighthouse}; and also two for reflectance and surface normal comparison~\cite{barron2013intrinsic,li2020inverse}. Lighthouse~\cite{srinivasan2020lighthouse} takes perspective stereo images as input to estimate the environment lighting given a position within the scene. Li~\etal~\cite{li2020inverse} takes a single image while Barron~\etal~\cite{barron2013intrinsic} takes a RGB-D input to estimate the reflectance, normal, depth, and lighting. 
All the above methods only take input with the resolution to be $240 \times 320$.
To satisfy their input requirement, we crop the middle region of our \threesixty stereo input into small patches with the target resolution. 
The top and bottom regions of the input are discarded to avoid the distortions and simulate the views in perspective projection. 
Then we feed the stereo images to Lighthouse~\cite{srinivasan2020lighthouse}; a single image to Li~\etal~\cite{li2020inverse}; and a single image with depth estimation to Barron~\etal~\cite{barron2013intrinsic}.

\section{Experiments}
Since there are no previous works that use a similar setup for this task, we only compare our work with those using perspective images. 
The comparisons here aim at demonstrating the strengths of \threesixty images over the perspective images in the tasks of lighting estimation and intrinsic decomposition of the scene properties. 
\subsection{Lighting Estimation}
In \fref{fig:barbershop_results}, we present the inserted mirror-objects at different locations within the scene.  
Our method  estimates the lighting with elaborate high-frequency details. Our inserted objects correctly reflect the changes of lighting among different locations,  demonstrating the spatially-coherency of our lighting model.
The two strengths above jointly contribute to the appealing reflection effects of a virtual mirror-object. 

We provide quantitative comparisons between our estimated illumination maps and previous methods~\cite{li2020inverse,srinivasan2020lighthouse}  in \fref{fig:classroom_lighting}. 
As a result of a simplified lighting model, Li~\cite{li2020inverse}'s illumination map only contains low-frequency information, leading a mirror-object looks diffuse.
Lighthouse~\cite{srinivasan2020lighthouse}, instead, based on a simplified model of the scene's geometry, can recover part of the scene in low definition. However,  due to the limited field of view of their perspective input, their hallucination on the unobserved scene is far from satisfactory. 
By the virtue of the \threesixty input, our \envlight provides an accurate representation of the lighting and geometry of the scene. 
Hence, our illumination maps are all estimated in high-definition and very close to the ground truth. 

\subsection{Reflectance and Normal Estimation}
\begin{figure}
	\centering
	\rotatebox{90}{ {\footnotesize \hspace{6mm} GT \hspace{17mm} Ours \hspace{12mm} Li~\etal~\cite{li2020inverse} \hspace{5.5mm} Barron~\etal~\cite{barron2013intrinsic} \hspace{8mm} Input}}\includegraphics[width=0.46\textwidth]{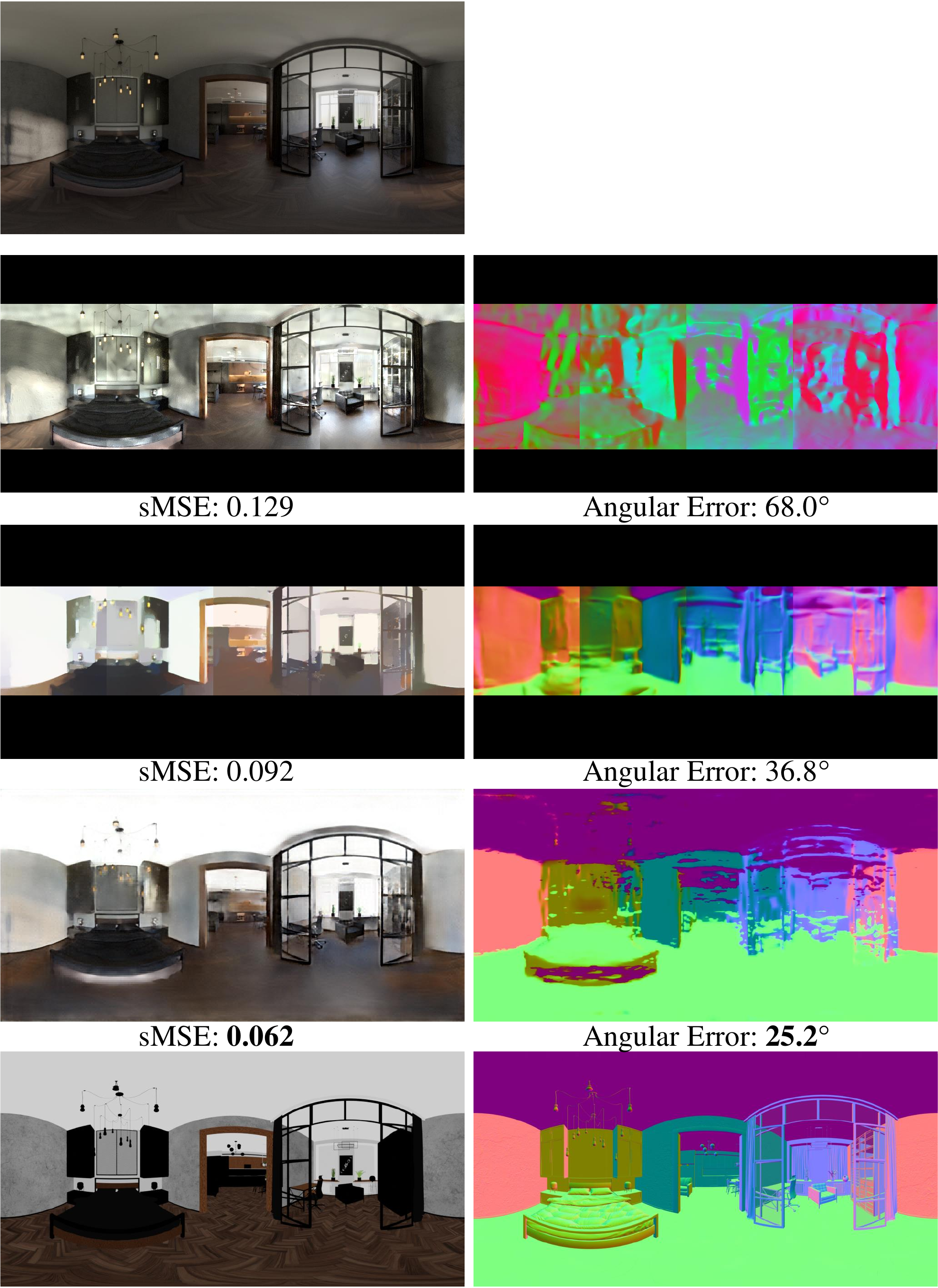}
	\caption{Estimated reflectance and normal on our synthetic scene `bedroom.' The sMSE and MAE are shown at the bottom.
    }
	\label{fig:albedo_normal}
\end{figure}
\begin{table}
\centering
{\small
\begin{tabular}{|c|c|c|}
\hline
Methods & Reflectance & Normal \\
\hline
Barron~\etal~\cite{barron2013intrinsic}  &   0.108   & 70.6          \\
\hline
Li~\etal~\cite{li2020inverse}   &  0.084   & 34.0 \\
\hline
Ours   &  \textbf{0.073}   & \textbf{24.6} \\
\hline
\end{tabular}
}
\caption{Quantitative comparison on our synthetic scenes. We use sMSE for reflectance and MAE in degrees for normal.}
\label{tab:albedo_normal}
\end{table}
We showcase the quality of our estimated reflectance map and surface normal map in the first row of \fref{fig:barbershop_results}. 
We also present quantitative results in \fref{fig:albedo_normal} and \Tref{tab:albedo_normal}. We use scale-invariant mean-square-error (sMSE) for reflectance and mean angular error (MAE) in degrees for normal.  Lower is better for both the metrics.
As mentioned in \sref{sec:data_eval},  all the competing methods~\cite{barron2013intrinsic,li2020inverse} take small resolution perspective input.  
To avoid the heavy distortion on \threesixty images,  we crop the middle region of our \threesixty image into four $240\times 320$ image patches as the input for others. Then we merge the results for viewing and comparison. The discarded regions are displayed in black in their results.
Although our method can estimate the entire scene, the quantitative evaluation shown in \fref{fig:albedo_normal} and \Tref{tab:albedo_normal} is only computed on those regions that are both generated by all the methods for a fair comparison.

Li~\etal~\cite{li2020inverse} estimates the reflectance from prior knowledge and past data. They present a wrong estimation on the `wall' region in \fref{fig:albedo_normal}, which is likely caused by the overfitting on a `white wall.' Our method takes the \threesixty full observation of the scene to jointly reason lighting, geometry, and reflectance by physical insights. Therefore, our estimation is of higher accuracy.

\subsection{Testing on Real Data} 
\begin{figure}
	\centering
	\includegraphics[width=0.48\textwidth]{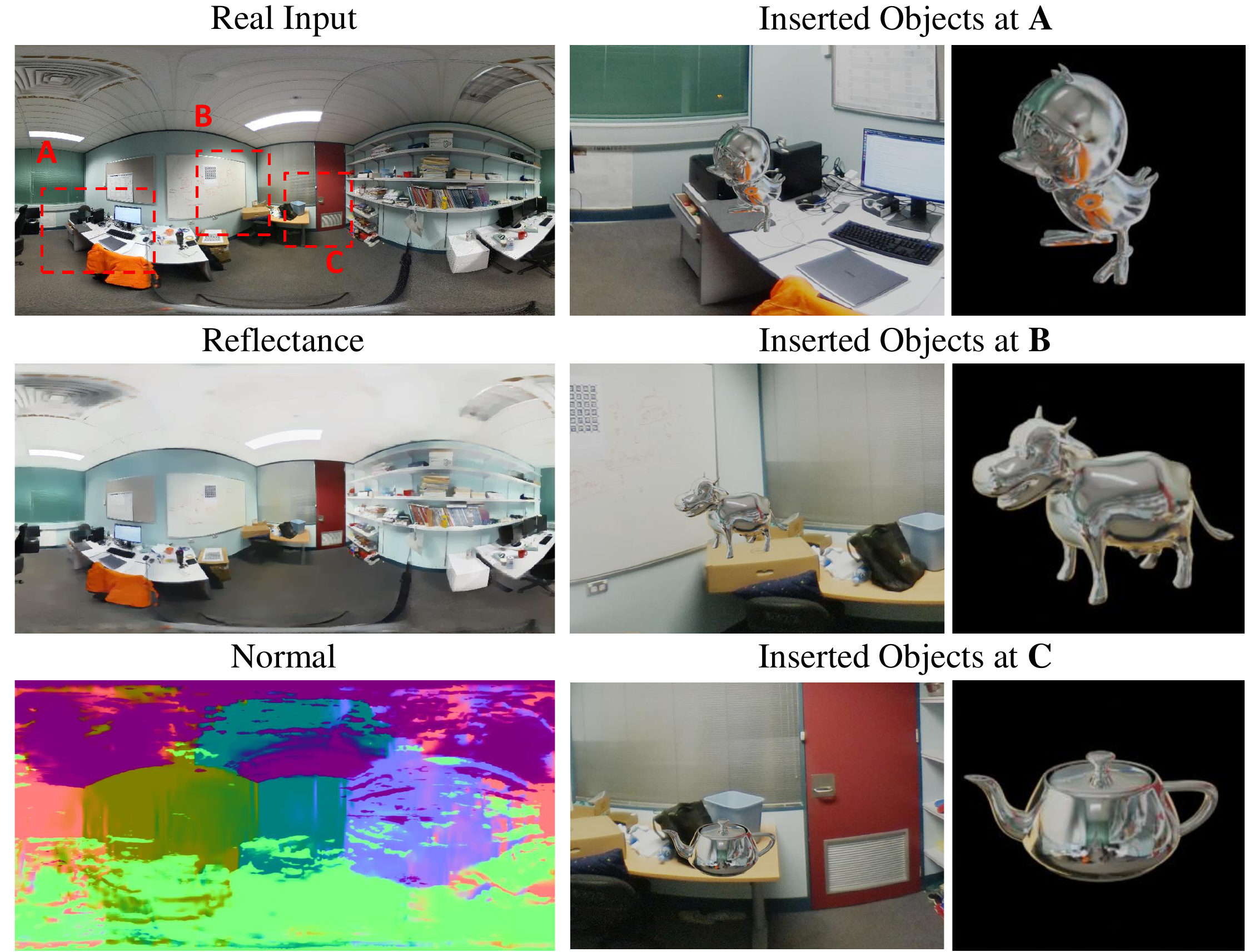}
	\caption{Results of real images. 
    Both the reflectance and normal are of high quality except the ceiling region.
    We relight three objects by our illumination map at different locations.
    Note how the three inserted mirror-objects correctly reflect the surroundings at each location: the orange cloth reflected on the `duck'; the ceiling light reflected on the `cow'; the scene reflected on the `teapot.' }
	\label{fig:office_real}
\end{figure}

Our results on the real data are shown in  \fref{fig:office_real}.  
The top-bottom camera setting has difficulties in capturing the ceiling and floor regions with accurate depth, as they are either occluded by the tripod or containing severe distortions that may lead to wrong disparities. 
Hence, our method fail to estimate the normal at the part of the ceiling region in this scene.
This kind of noise hardly occurs in the synthetic data, which does not have the occlusion problem and noises. 

\subsection{Ablation Study}

\begin{table}
\centering
{\small
\begin{tabular}{|c|c|c|c|}
\hline
    &  $\Phi_1$   &  $\Phi_{1+\frac{1}{4}}$ & Render and Refine \\
\hline
Reflectance  &   0.086   & 0.080   & \textbf{0.073}       \\
\hline
Normal &  47.3   & \textbf{24.6}  &  -\\
\hline
\end{tabular}
}
\caption{Quantitative results on ablated versions of our model.  We use sMSE for reflectance and MAE for normal.}
\label{tab:ablation}
\end{table}

\Tref{tab:ablation} shows the quantitative results on the ablated versions of our method.
From left to right, the columns denote the origin RN-Net, pyramid RN-Net, and our full method with the rendering and total variation refinement, respectively. 
It shows that the pyramid structure improves reflectance and normal. We also observe that the rendering and refinement module  can effectively reduce the noise and outliers to provide more plausible reflectance maps.

\section{Conclusion}
We have presented a method that takes a \threesixty panoramic stereo as input, and jointly estimates spatially-varying and 3D coherent lighting in high-definition, reflectance, and geometry of the entire scene. Instead of using a regular camera with a limited field of view, we demonstrate  the advantages of \threesixty input in observing and estimating the whole scene. Our lighting model accurately reconstructs 3D illumination maps, enabling mirror-like objects to be inserted in the scene with realistic effect. We also leverage the physical constraints between the lighting and geometry to infer both surface reflectances and normals of the environment. Our results outperform previous state-of-the-art, both quantitatively and qualitatively. Results on synthetic and real images confirm the effectiveness and practicability of our method, by a simple \threesixty stereo setup.

\textbf{Acknowledgement.}
This research is funded in part by the ARC Centre of Excellence for Robotics Vision (CE140100016) and ARC-Discovery (DP 190102261).
Yasuyuki Matsushita is supported by JSPS KAKENHI Grant Number JP19H01123.

\newpage
{\small
\bibliographystyle{ieee_fullname}
\bibliography{egbib}

\begin{thebibliography}{10}\itemsep=-1pt

\bibitem{banterle2013envydepth}
Francesco Banterle, Marco Callieri, Matteo Dellepiane, Massimiliano Corsini,
  Fabio Pellacini, and Roberto Scopigno.
\newblock Envydepth: An interface for recovering local natural illumination
  from environment maps.
\newblock In {\em Computer Graphics Forum}, volume~32, pages 411--420. Wiley
  Online Library, 2013.

\bibitem{barron2013intrinsic}
Jonathan~T Barron and Jitendra Malik.
\newblock Intrinsic scene properties from a single rgb-d image.
\newblock In {\em Proceedings of the IEEE Conference on Computer Vision and
  Pattern Recognition}, pages 17--24, 2013.

\bibitem{barron2014shape}
Jonathan~T Barron and Jitendra Malik.
\newblock Shape, illumination, and reflectance from shading.
\newblock {\em IEEE transactions on pattern analysis and machine intelligence},
  37(8):1670--1687, 2014.

\bibitem{bell2014intrinsic}
Sean Bell, Kavita Bala, and Noah Snavely.
\newblock Intrinsic images in the wild.
\newblock {\em ACM Transactions on Graphics (TOG)}, 33(4):1--12, 2014.

\bibitem{bi20151}
Sai Bi, Xiaoguang Han, and Yizhou Yu.
\newblock An l 1 image transform for edge-preserving smoothing and scene-level
  intrinsic decomposition.
\newblock {\em ACM Transactions on Graphics (TOG)}, 34(4):1--12, 2015.

\bibitem{blender}
Blender~Online Community.
\newblock Blender - a 3d modelling and rendering package, 2020.

\bibitem{debevec2008rendering}
Paul Debevec.
\newblock Rendering synthetic objects into real scenes: Bridging traditional
  and image-based graphics with global illumination and high dynamic range
  photography.
\newblock In {\em ACM SIGGRAPH 2008 classes}, pages 1--10. 2008.

\bibitem{fan2018revisiting}
Qingnan Fan, Jiaolong Yang, Gang Hua, Baoquan Chen, and David Wipf.
\newblock Revisiting deep intrinsic image decompositions.
\newblock In {\em Proceedings of the IEEE conference on computer vision and
  pattern recognition}, pages 8944--8952, 2018.

\bibitem{gardner2019deep}
Marc-Andr{\'e} Gardner, Yannick Hold-Geoffroy, Kalyan Sunkavalli, Christian
  Gagn{\'e}, and Jean-Francois Lalonde.
\newblock Deep parametric indoor lighting estimation.
\newblock In {\em 2019 IEEE/CVF International Conference on Computer Vision
  (ICCV)}, pages 7174--7182. IEEE.

\bibitem{gardner2017learning}
Marc-Andr{\'e} Gardner, Kalyan Sunkavalli, Ersin Yumer, Xiaohui Shen, Emiliano
  Gambaretto, Christian Gagn{\'e}, and Jean-Fran{\c{c}}ois Lalonde.
\newblock Learning to predict indoor illumination from a single image.
\newblock {\em ACM Transactions on Graphics (TOG)}, 36(6):1--14, 2017.

\bibitem{garon2019fast}
Mathieu Garon, Kalyan Sunkavalli, Sunil Hadap, Nathan Carr, and
  Jean-Fran{\c{c}}ois Lalonde.
\newblock Fast spatially-varying indoor lighting estimation.
\newblock In {\em Proceedings of the IEEE Conference on Computer Vision and
  Pattern Recognition}, pages 6908--6917, 2019.

\bibitem{he2016deep}
Kaiming He, Xiangyu Zhang, Shaoqing Ren, and Jian Sun.
\newblock Deep residual learning for image recognition.
\newblock In {\em Proceedings of the IEEE conference on computer vision and
  pattern recognition}, pages 770--778, 2016.

\bibitem{im2016all}
Sunghoon Im, Hyowon Ha, Fran{\c{c}}ois Rameau, Hae-Gon Jeon, Gyeongmin Choe,
  and In~So Kweon.
\newblock All-around depth from small motion with a spherical panoramic camera.
\newblock In {\em European Conference on Computer Vision}, pages 156--172.
  Springer, 2016.

\bibitem{kim20133d}
Hansung Kim and Adrian Hilton.
\newblock 3d scene reconstruction from multiple spherical stereo pairs.
\newblock {\em International journal of computer vision}, 104(1):94--116, 2013.

\bibitem{kingma2014adam}
Diederik~P Kingma and Jimmy Ba.
\newblock Adam: A method for stochastic optimization.
\newblock {\em arXiv preprint arXiv:1412.6980}, 2014.

\bibitem{legendre2019deeplight}
Chloe LeGendre, Wan-Chun Ma, Graham Fyffe, John Flynn, Laurent Charbonnel, Jay
  Busch, and Paul Debevec.
\newblock Deeplight: Learning illumination for unconstrained mobile mixed
  reality.
\newblock In {\em Proceedings of the IEEE Conference on Computer Vision and
  Pattern Recognition}, pages 5918--5928, 2019.

\bibitem{li2004stereo}
Yin Li, Heung-Yeung Shum, Chi-Keung Tang, and Richard Szeliski.
\newblock Stereo reconstruction from multiperspective panoramas.
\newblock {\em IEEE Transactions on Pattern Analysis and Machine Intelligence},
  26(1):45--62, 2004.

\bibitem{li2020inverse}
Zhengqin Li, Mohammad Shafiei, Ravi Ramamoorthi, Kalyan Sunkavalli, and
  Manmohan Chandraker.
\newblock Inverse rendering for complex indoor scenes: Shape, spatially-varying
  lighting and svbrdf from a single image.
\newblock In {\em Proceedings of the IEEE/CVF Conference on Computer Vision and
  Pattern Recognition}, pages 2475--2484, 2020.

\bibitem{li2018cgintrinsics}
Zhengqi Li and Noah Snavely.
\newblock Cgintrinsics: Better intrinsic image decomposition through
  physically-based rendering.
\newblock In {\em Proceedings of the European Conference on Computer Vision
  (ECCV)}, pages 371--387, 2018.

\bibitem{lombardi2015reflectance}
Stephen Lombardi and Ko Nishino.
\newblock Reflectance and illumination recovery in the wild.
\newblock {\em IEEE transactions on pattern analysis and machine intelligence},
  38(1):129--141, 2015.

\bibitem{narihira2015direct}
Takuya Narihira, Michael Maire, and Stella~X Yu.
\newblock Direct intrinsics: Learning albedo-shading decomposition by
  convolutional regression.
\newblock In {\em Proceedings of the IEEE international conference on computer
  vision}, pages 2992--2992, 2015.

\bibitem{rudin1992nonlinear}
Leonid~I Rudin, Stanley Osher, and Emad Fatemi.
\newblock Nonlinear total variation based noise removal algorithms.
\newblock {\em Physica D: nonlinear phenomena}, 60(1-4):259--268, 1992.

\bibitem{shi2017learning}
Jian Shi, Yue Dong, Hao Su, and Stella~X Yu.
\newblock Learning non-lambertian object intrinsics across shapenet categories.
\newblock In {\em Proceedings of the IEEE Conference on Computer Vision and
  Pattern Recognition}, pages 1685--1694, 2017.

\bibitem{song2019neural}
Shuran Song and Thomas Funkhouser.
\newblock Neural illumination: Lighting prediction for indoor environments.
\newblock In {\em Proceedings of the IEEE Conference on Computer Vision and
  Pattern Recognition}, pages 6918--6926, 2019.

\bibitem{srinivasan2020lighthouse}
Pratul~P Srinivasan, Ben Mildenhall, Matthew Tancik, Jonathan~T Barron, Richard
  Tucker, and Noah Snavely.
\newblock Lighthouse: Predicting lighting volumes for spatially-coherent
  illumination.
\newblock In {\em Proceedings of the IEEE/CVF Conference on Computer Vision and
  Pattern Recognition}, pages 8080--8089, 2020.

\bibitem{wang2020360sd}
Ning-Hsu Wang, Bolivar Solarte, Yi-Hsuan Tsai, Wei-Chen Chiu, and Min Sun.
\newblock 360sd-net: 360 stereo depth estimation with learnable cost volume.
\newblock In {\em 2020 IEEE International Conference on Robotics and Automation
  (ICRA)}, pages 582--588. IEEE, 2020.

\bibitem{xing2018automatic}
Guanyu Xing, Yanli Liu, Haibin Ling, Xavier Granier, and Yanci Zhang.
\newblock Automatic spatially varying illumination recovery of indoor scenes
  based on a single rgb-d image.
\newblock {\em IEEE Transactions on Visualization and Computer Graphics}, 2018.

\bibitem{zhao2020pointar}
Yiqin Zhao and Tian Guo.
\newblock Pointar: Efficient lighting estimation for mobile augmented reality.
\newblock {\em arXiv preprint arXiv:2004.00006}, 2020.

\bibitem{zheng2019structured3d}
Jia Zheng, Junfei Zhang, Jing Li, Rui Tang, Shenghua Gao, and Zihan Zhou.
\newblock Structured3d: A large photo-realistic dataset for structured 3d
  modeling.
\newblock {\em arXiv preprint arXiv:1908.00222}, 2019.

\bibitem{zhou2019glosh}
Hao Zhou, Xiang Yu, and David Jacobs.
\newblock Glosh: Global-local spherical harmonics for intrinsic image
  decomposition.
\newblock In {\em 2019 IEEE/CVF International Conference on Computer Vision
  (ICCV)}, pages 7819--7828. IEEE.

\end{thebibliography}
}

\end{document}